\documentclass[
]{ceurart}

\usepackage{listings}
\usepackage{xcolor}
\usepackage{float}

\begin{document}

\copyrightyear{2024}
\copyrightclause{Copyright for this paper by its authors.
  Use permitted under Creative Commons License Attribution 4.0
  International (CC BY 4.0).}

\conference{CHR 2024: Computational Humanities Research Conference, December 4–6, 2024, Aarhus, Denmark}

\title{Annotating References to Mythological Entities in French Literature}
\subtitle{Assessing the Strengths and Limitations of ChatGPT (and Other Large Language Models) for the Task}

\author{Thierry Poibeau}[
orcid={0000-0003-3669-4051},
email=thierry.poibeau@ens.psl.eu,
url=https://prairie-institute.fr/chairs/poibeau-thierry/
]
\address{Laboratoire Lattice\\CNRS \& ENS-PSL \& U. Sorbonne nouvelle\\
1, rue Maurice Arnoux, 92120 Montrouge, France}


\begin{abstract}
  In this paper, we explore the relevance of large language models (LLMs) for annotating references to Roman and Greek mythological entities in modern and contemporary French literature. We present an annotation scheme and demonstrate that recent LLMs can be directly applied to follow this scheme effectively, although not without occasionally making significant analytical errors. Additionally, we show that LLMs (and, more specifically, ChatGPT) are capable of offering interpretative insights into the use of mythological references by literary authors. However, we also find that LLMs struggle to accurately identify relevant passages in novels (when used as an information retrieval engine), often hallucinating and generating fabricated examples—an issue that raises significant ethical concerns. Nonetheless, when used carefully, LLMs remain valuable tools for performing annotations with high accuracy, especially for tasks that would be difficult to annotate comprehensively on a large scale through manual methods alone.
  \end{abstract}


\maketitle

\section{Introduction}

Mythological allusions often serve as powerful tools for authors, offering symbolic depth and universal themes that resonate across cultures and time. They enrich the storytelling by connecting contemporary works to timeless archetypes and motifs. For example, in the following passage coming from Marcel Proust’s \textit{In Search of Lost Time}, the author reflects on the dissolution of mythological wonder in the modern world:

\begin{quotation}
\noindent
\textit{Les géographes, les archéologues nous conduisent bien dans l’île de Calypso, exhument bien le palais de Minos. Seulement Calypso n’est plus qu’une femme, Minos, qu’un roi sans rien de divin. Ainsi s’était dissipée toute la gracieuse mythologie océanique que J’avais composée les premiers jours.} \\\textit{In the Shadow of Young Girls in Flower} (\textit{À l'ombre des jeunes filles en fleurs}, also translated as \textit{Within a Budding Grove}), 1919.\footnote{\textit{Geographers and archaeologists do indeed lead us to the island of Calypso and unearth the palace of Minos. Yet, Calypso is now merely a woman, and Minos, a king devoid of any divinity. Thus, the graceful oceanic mythology I had imagined in the early days had entirely dissipated. }}
\end{quotation}

\noindent
Here, Proust uses mythological references to highlight the loss of enchantment and the reduction of myths to mundane realities.

	
Today, we have access to vast collections of digitized corpora. For French literature, Gallica provides more than 15,000 novels, most with high-quality OCR \cite{langlais2021}, making these corpora suitable for computational exploration, commonly referred to as cultural analytics. Furthermore, specialized annotation frameworks like BookNLP \cite{github_bamman} have been developed, enabling the analysis of entities (such as characters and locations \cite{bamman-2019-annotated}) and coreference resolution (linking all mentions of the same real-world entity \cite{bamman-etal-2020-annotated,poesio2023computational}).

However, there are known challenges with such annotation frameworks \cite{mcgillivray2020}. For instance, BookNLP uses the tag “PER” for persons/characters, but how should we categorize speaking animals or Gods that influence the narrative? While frameworks like BookNLP offer a basic, generic annotation layer useful for many studies, researchers often require additional, more specialized layers with finer-grained categories. These custom annotation schemes may also need to reference external ontologies, introducing the need for entity linking \cite{shen2014} (a feature not provided by the original BookNLP) or even necessitating the creation of custom tools.

Consider, for example, the case of Greek mythology. Our goal is to identify references to Greek mythology in 19th-century French literature, with the hypothesis that many of these references are metaphorical or used to highlight a specific trait of a novel’s protagonist. In this context, the focus is not only (and perhaps not primarily) on identifying the Gods and other mythological figures themselves, but rather on understanding how these references relate to the qualities, mindset, or symbolic traits of another character in the narrative.


This raises several important research questions: How can mythological references be effectively recognized in contemporary novels? How can different annotation levels and schemes be combined within the same text? How can we annotate the symbolic meaning or power of mythological references? And finally, what resources and tools would be best suited for these tasks?

This paper is structured as follows. We first show that LLMs (and, more specifically, ChatGPT) can be used effectively to annotate and interpret relevant mentions to mythological entities in texts, although not without occasionally making significant analytical errors. We then  observe that the same models, somewhat unexpectedly, struggle when used as retrieval engines in the same context. The paper finishes on a discussion and some perspectives based on our observations. 

\section{Previous Work}

The study of mythological entity annotation in recent literature has garnered increasing interest within the digital humanities, particularly as natural language processing (NLP) tools advance. While much focus in early research has been on classical texts, such as Greek and Roman literature, recent efforts have shifted towards identifying mythological references in contemporary texts, especially novels. The goal is not only to detect these entities but also to analyze their metaphorical or symbolic usage, especially in contexts where references may reflect broader cultural or narrative functions \cite{graf1993}.

Several existing resources aid in the annotation of mythological entities. The Theoi Project\footnote{Theoi Project: \url{https://www.theoi.com/}} is a foundational reference for Greek mythology, offering structured information on Gods, heroes, and mythical creatures. While not explicitly designed for computational tasks, this resource is widely used for manual annotation and can be adapted for automatic processing when combined with other tools. Additionally, the Perseus Digital Library\footnote{Perseus Digital Library: \url{https://www.perseus.tufts.edu/hopper/}} provides extensive texts and annotations from classical antiquity, though its focus remains primarily on ancient languages, limiting its direct applicability to modern literature.

More specialized efforts in computational mythology include the development of the Ontology of Greek Mythology (OGM) \cite{pastor2021}, which builds on Wikidata and integrates data from sources like the Theoi Project. This ontology is a structured knowledge graph that captures complex relationships among mythological figures, locations, and events. OGM represents not only the mythological entities themselves but also accommodates multiple narrative variations, making it suitable for comparative mythology studies. Its adaptability offers promising opportunities for studying references in modern literature where mythological themes often take on new metaphorical meanings.

In terms of NLP tools, BookNLP, developed by David Bamman at Berkeley, offers significant potential for annotating novels \cite{github_bamman,bamman-2019-annotated,bamman-etal-2020-annotated}. Originally created to recognize characters, events, and quotations in English novels, it has since been adapted for other languages, including French \cite{github_lattice,melanie2024,barre_operationalizing_2023}. BookNLP can be useful for identifying specific entities when combined with external resources such as ontologies. However, it lacks specific models for mythology-focused annotation, requiring either customization or integration with other tools to accurately detect references to figures like Zeus or Athena in modern texts.



\section{An Annotation Scheme for Mythological Entities}

Our annotation scheme for annotating mythological entities in contemporary novels was created to address the complexity and variety of mythological references in modern literature. Existing tagsets for entity recognition, such as those used for general Named Entity Recognition (NER) tasks, do not capture the specific nuances of mythological references, which can include not only figures like Gods and creatures but also abstract concepts, locations, and significant objects from myths. By defining tags like “deity,” “hero,” “event,” and “concept,” this schema allows for a richer and more precise annotation process.

The tagset builds on previous work in computational mythology, such as the use of ontologies like the Ontology of Greek Mythology (OGM), but extends it to fit the context of contemlporary novels where mythological references often serve metaphorical or symbolic purposes. Here is an overview of the categories considered.  

\begin{itemize}
\item  \color{blue}<mythEntity type="deity">: \color{black}For gods or goddesses, like Zeus or Athena.
\item  \color{blue}<mythEntity type="hero">: \color{black}For mythological heroes like Heracles or Achilles.
\item  \color{blue}<mythEntity type="creature">: \color{black}For mythological creatures like the Minotaur.
\item  \color{blue}<mythEntity type="half\_creature">: \color{black}For beings like centaurs or satyrs.
\item  \color{blue}<mythEntity type="creature\_group">: \color{black}For groups of mythological entities like the Gorgons or collective references to “monsters.”
\item  \color{blue}<mythEntity type="monsters">: \color{black}For general references to mythological beings or monsters.
\item  \color{blue}<mythEntity type="location">: \color{black}For places or geographic references like Érymanthos.
\item  \color{blue}<mythEntity type="event">: \color{black}For mythological events, like the Trojan War or the Labors of Heracles.
\item  \color{blue}<mythEntity type="object">: \color{black}For mythological objects, like the Golden Fleece or Pandora’s Box.
\item  \color{blue}<mythEntity type="concept">: \color{black}For abstract mythological concepts, like Fate or Nemesis.
\end{itemize}

\noindent
This tagset can help structure and clarify mythological references, distinguishing between individual creatures, groups, and locations. It can easily be expanded, for instance, to specify which mythology the entity belongs to (Greek, Roman, or other), or to include a more refined categorization of events (to include classes such as community-centric, character-centric, environmental, etc.).

\section{Entity Annotation with LLMs}

As discussed, some resources exist for annotating mythological references in novels, but it appears that they are not as effective as large language models (LLMs). These resources are often limited by language-specific variations (such as names of Gods differing across languages), are not always comprehensive, and may provide irrelevant or excessive information that complicates the annotation process.

In contrast, LLMs, which are trained on extensive datasets including mythological information from sources like Wikipedia, offer a more powerful and adaptable solution for mythological entity annotation. In our tests with Mistral, LLaMA3 and ChatGPT (version 3 and 4o), more than 95\% of mythological entities were recognized. These models demonstrate impressive performance by drawing from their broad training data and applying it effectively across different contexts.

Below is the example from Marcel Proust, already cited above, annotated by ChatGPT (version 4o) using the annotation scheme presented in the previous section (see \textbf{Annex 1} for details on the prompt we used for annotation). This demonstrates how LLMs can identify mythological references in a literary context,  although with numerous errors in the type and span of entities to annotate.

\begin{quotation}
\noindent
\textit{Les géographes, les archéologues nous conduisent bien dans l’île de \color{blue}<mythEntity type="location">\color{olive}Calypso<\color{blue}/mythEntity>\color{black}, exhument bien le palais de \color{blue}<mythEntity type= "deity">\color{olive}Minos\color{blue}</mythEntity>.\color{black} Seulement \color{blue}<mythEntity type="location">\color{olive}Calypso\color{blue}</mythEntity>\color{black} n’est plus qu’une femme, \color{blue}<mythEntity type= "deity">\color{olive}Minos<\color{blue}/mythEntity>\color{black}, qu’un roi sans rien de divin. Ainsi s’était dissipée toute la gracieuse \color{blue}<mythEntity type="concept">\color{olive}mythologie océanique\color{blue}</mythEntity>\color{black} que J’avais composée les premiers jours.}
\end{quotation}

\noindent
The first occurrence is correctly recognized as a location name, but it is “île de Calypso” that should be annotated, not just “Calypso.” Then "Minos" is identified as a deity, whereas it is “le Palais de Minos” that should be recognized and analyzed as a location name. Then, Calypso is again identified as a location name, even though the text at this point clearly refers to the deity. This illustrates a “contagion” phenomenon, where different occurrences tend to be assigned the same type, despite distinct and context-specific indications. In subsequent interactions, we also observed that ChatGPT can sometimes alter the original text (e.g., replacing the second occurrence of “Calypso” with “île de Calypso” to align it with the first occurrence). This highlights the importance of using ChatGPT with caution and carefully verifying that the original text remains unaltered.

Below is another example (all the examples are from Marcel Proust’s \textit{In Search of Lost Time}). Annotation proves more reliable with isolated examples, suggesting that the previous issue was primarily caused by the varying contexts in which the same entity appeared, as said above.

\begin{quotation}
\noindent
Si alors passait la princesse de Guermantes, belle et
légère comme \color{blue}<mythEntity type="deity">\color{olive}Diane\color{blue}</mythEntity>\color{black}, laissant traîner derrière elle
un manteau incomparable, faisant se détourner
toutes les têtes et suivie par tous les yeux (par
ceux de Mme de Cambremer plus que par tous les
autres), M. de Beausergent s’absorbait dans une
conversation avec sa voisine
\end{quotation}

\noindent
And finally, another example related to Roman mythology (\textit{Minerve}).

\begin{quotation}
\noindent
Je ne pensais pas qu’aucune femme pût usurper le corsage pailleté de l’autre plus que \color{blue}<mythEntity type="object">\color{olive}l’égide étincelante et frangée\color{blue}</mythEntity>\color{black}de \color{blue}<mythEntity type="deity">\color{olive}Minerve\color{blue}</mythEntity>\color{black}.
\end{quotation}

These are just examples. Our study is not comprehensive here, but this shows that  it possible for a system to annotate pretty accurately French novels including sporadic, albeit real, references to Roman or Greek mythology. 

More precisely, our experiments on 200 examples from 19th-century French novels (chosen randomly from the Gallica collection) using ChatGPT-4o demonstrate that 98\% of the entities were recognized, with more than  95\% being accurately annotated. The majority of issues encountered were consistent with the challenges previously described, such as incorrect spans or misclassification of entity types. These errors were often attributable to ambiguous or conflicting local contexts, which posed difficulties for the model in accurately interpreting and annotating specific occurrences. Note that these errors remain, however, marginal, and our observations indicate that ChatGPT can be effectively used to annotate texts, particularly when the process includes a phase of manual review and correction.

\section{Interpreting References to Mythological Entities in French Novels}

However, our primary goal is not just to recognize reference to mythological entities, but to analyze their contribution to the text \cite{proust-mythes}. Mythological references can offer cultural or historical commentary. By invoking ancient myths, authors connect the present with the past, highlighting continuities or contrasts in human behavior and societal values. In contemporary French novels, for instance, references to Greek Gods or mythological events might serve to critique contemporary issues or explore human psychology in ways that transcend time and place \cite{crant76}. This intertextuality (or transtextuality, as Genette calls it \cite{genette1992}) enriches the narrative and invites readers to engage with multiple layers of interpretation. Through myth, authors can address complex emotions, ideals, and conflicts with a shorthand that draws on collective cultural knowledge \cite{kristeva1969}.

To what extent can this kind of interpretation be deduced by LLMs from the original text? ChatGPT (4o) can produce the following comment from the first passage given as an example above: 

\begin{quotation}
``\textit{In this passage, Proust reflects on the disenchantment of the modern world, where the mythological and divine have been reduced to mundane historical facts. By stating that geographers and archaeologists have found the island of Calypso and the palace of Minos, he suggests that these once-magical places are now simply physical locations devoid of their former mythic grandeur. Calypso is no longer a goddess, and Minos is just a king, stripped of their divine and symbolic power. Proust is mourning the loss of the imaginative, poetic dimension that mythologies once offered.}''  
\end{quotation}

\noindent
This interpretation is correct, but probably reflect sources directly available on the Web. 

Care must be taken to ensure that the language model is not hallucinating or creating spurious interpretations. This is especially important for books whose content may not be part of the model’s training data (such as most of those from Gallica, which should not be used without prior agreement for training commercial models) \cite{chang-etal-2023-speak}. Although, in our case, interpretations rely on widely known information available in reference databases, particularly Wikipedia, as previously mentioned, this does not completely eliminate the risk of hallucination. LLMs can be considered as valuable tools for providing immediate interpretations of local contexts, such as when a character is directly compared to a mythological figure or when a mythological entity is invoked in relation to a specific character.

\section{The Inability of LLMs to Retrieve Relevant Passages and the Risk of Hallucination}

In the previous sections, we showed that large language models (LLMs), such as ChatGPT, LLaMA, and Mistral, can accurately annotate references to mythology. Their analysis tends to be correct, often more thorough than manual annotations, as they can easily detect rare references that may otherwise be missed across lengthy texts. But are LLMs equally good at retrieving passages containing references to mythological entities or events?

Hallucination is a well-documented issue with large language models (\cite{huang2024}. In our experiments,  this phenomenon was observed, particularly when the system was asked to retrieve specific passages rather than merely annotating them. For instance, when asked to extract passages where an author explicitly references Greek mythology, ChatGPT was most of the time generating convincing but entirely fabricated examples\footnote{In this experiment, we instructed the system to retrieve passages from a specific novel without providing the full text to ChatGPT. We regard this as the default way to use ChatGPT in this context.}. While we refrain from reproducing such instances here to avoid spreading forged texts (see however \textbf{Annex 2} for screenshots reflecting these hallucinations), this behavior appears to be quite prevalent as of October 2024—more the rule than the exception (this was still the case when revising this paper, in December 2024). ChatGPT often fails to acknowledge when no relevant passage is found and may even fabricate examples, even in cases where valid passages do exist but are not proposed as answers.

This risk can be especially troubling in educational settings, where such tools could be used for literary analysis. Even a tool like NotebookLM\footnote{NotebookLM: \url{https://notebooklm.google.com/}}, designed to process more accurately a predefined set of texts, can exhibit similar behavior. While NotebookLM is generally more reliable and capable of responding with “no relevant passage found,”\footnote{For example: ``Il n'y a aucune mention de la mythologie grecque dans les extraits de "Les Misérables" de Victor Hugo que vous avez fournis.'' (“There is no mention of Greek mythology in the excerpts from ‘Les Misérables’ by Victor Hugo that you provided.”).} it eventually resorts to generating false examples (after a few similar queries), possibly to avoid giving too many inconclusive answers.

At the time of writing the first version of this paper (October 2024), this behavior represented the default operational mode of the system. At the time of revising this paper (December 2024), however, NotebookLM demonstrated increased reliability and a reduced tendency toward hallucinations. Nevertheless, a new issue emerged regarding the model’s handling of textual extracts. While the system retains the ability to track extracts, it occasionally exhibits confusion, which introduces additional challenges. Notably, the model may erroneously assert the absence of relevant examples within a text, even when such examples are clearly present (see \textbf{Annex 3}). In this regard, NotebookLM appears to prioritize silence (exhibiting low recall) over the generation of hallucinated content.
This issue is a serious concern and should be factored into ongoing discussions about the potential and limitations of LLMs in cultural analytics and education. It also highlights the rapid evolution of these models, where an observation made on one day may no longer hold true the next. This raises significant concerns regarding their robustness and the reproducibility of experiments involving LLMs.

One might wonder why hallucinations appear in the passage retrieval context but not in the annotation context. This may seem counterintuitive, as both tasks are closely related and seem to require similar analysis. However, the type of task clearly plays a major role, likely due to adjustments made by developers to the chat function of the system, especially through the Reinforcement Learning from Human Feedback phase (RLHF) \cite{kaufmann2024surveyreinforcementlearninghuman}, as opposed to the more generic GPT model that do not include RLHF. Annotation involves identifying isolated elements in the text and tagging them, where the system should not generate new content (beyond the annotations themselves).

In the case of a query intended to retrieve a passage, the system is designed (through the 'system prompt'\footnote{The ``system prompt'' in a system like ChatGPT refers to the initial instruction or configuration provided to the model to guide its behavior throughout the interaction. It establishes the context, tone, and constraints for the model’s responses, effectively shaping how it interprets user inputs and generates outputs.}) to produce a relevant answer by generating new text based on the user prompt (typically, a response to a question). In our specific case, this should normally involve retrieving a sentence or passage. However, unexpectedly, the system (likely in an effort to avoid plagiarism) creates an excerpt “in the style of” Proust, but not authentically from Proust. While this technical explanation may seem reasonable, it is highly problematic that the system produces fabricated quotations, even when explicitly instructed that quotations must be exact and directly sourced from the text under study. Further analysis is required to understand the underlying mechanics of the system’s failure in this case, that will complement broader investigations into the capabilities of large language models for cultural analytics \cite{chang-etal-2023-speak}. 

\section{Conclusion and Perspectives}

Expanding the annotation and analysis of mythological references to a large corpus of 15,000 French novels will open new opportunities for both quantitative and qualitative research. We can expect to uncover patterns in how mythological entities are invoked across different time periods, genres, and authors, revealing how modern French literature repurposes ancient myths. This broader dataset will allow for the identification of recurrent themes, metaphors, and cultural shifts in the treatment of mythology, offering insight into how these references shape characters and narratives over time.

This expanded analysis will also make it possible to trace the evolution of mythological archetypes and their transformation in different literary contexts. We can expect to conduct comparative studies between individual authors and movements, evaluating whether particular myths are associated with specific literary genres or thematic concerns (e.g., existentialism, romanticism, realism). Additionally, this large-scale study will enable the exploration of metaphorical uses of mythology, helping to determine how certain deities or myths become allegorical references for contemporary issues, such as power, identity, or morality.

\bibliography{bibliography}

\section{Annexes}

\subsection{Annex 1}

This is the prompt used for annotation (cf. section 4). We intentionally used simple prompts and avoided over-optimizing the strategy with advanced prompt engineering, as our aim was to evaluate ChatGPT’s capabilities for users with general expertise, rather than specialists in prompt engineering.

{\footnotesize 
\begin{verbatim}
This is the annotation schema: 
<mythEntity type="deity">: For gods or goddesses, like Zeus or Athena.
<mythEntity type="hero">: For mythological heroes like Heracles or Achilles.
<mythEntity type="creature">: For mythological creatures like the Minotaur.
<mythEntity type="half\_creature">: For beings like centaurs or satyrs.
<mythEntity type="creature\_group">: For groups of mythological entities like the Gorgons 
or collective references to “monsters.”
<mythEntity type="monsters">: For general references to mythological beings or monsters.
<mythEntity type="location">: For places or geographic references like Érymanthos.
<mythEntity type="event">: For mythological events, like the Trojan War or the Labors 
of Heracles.
<mythEntity type="object">: For mythological objects, like the Golden Fleece or Pandora’s Box.
<mythEntity type="concept">: For abstract mythological concepts, like Fate or Nemesis.
Please annotate the following sentence using this schema -> Les géographes, les 
archéologues nous conduisent bien dans l’île de Calypso, exhument bien le palais de Minos. 
Seulement Calypso n’est plus qu’une femme, Minos, qu’un roi sans rien de divin. Ainsi 
s’était dissipée toute la gracieuse mythologie océanique que J’avais composée les 
premiers jours. 
\end{verbatim}

\subsection{Annex 2}

In this section, we provide examples of instances where ChatGPT generates hallucinated content while attempting to retrieve passages from specific novels that reference mythology  (\textbf{Figure 1}).

\begin{figure}[ht]
\centering
\includegraphics[scale=.55]{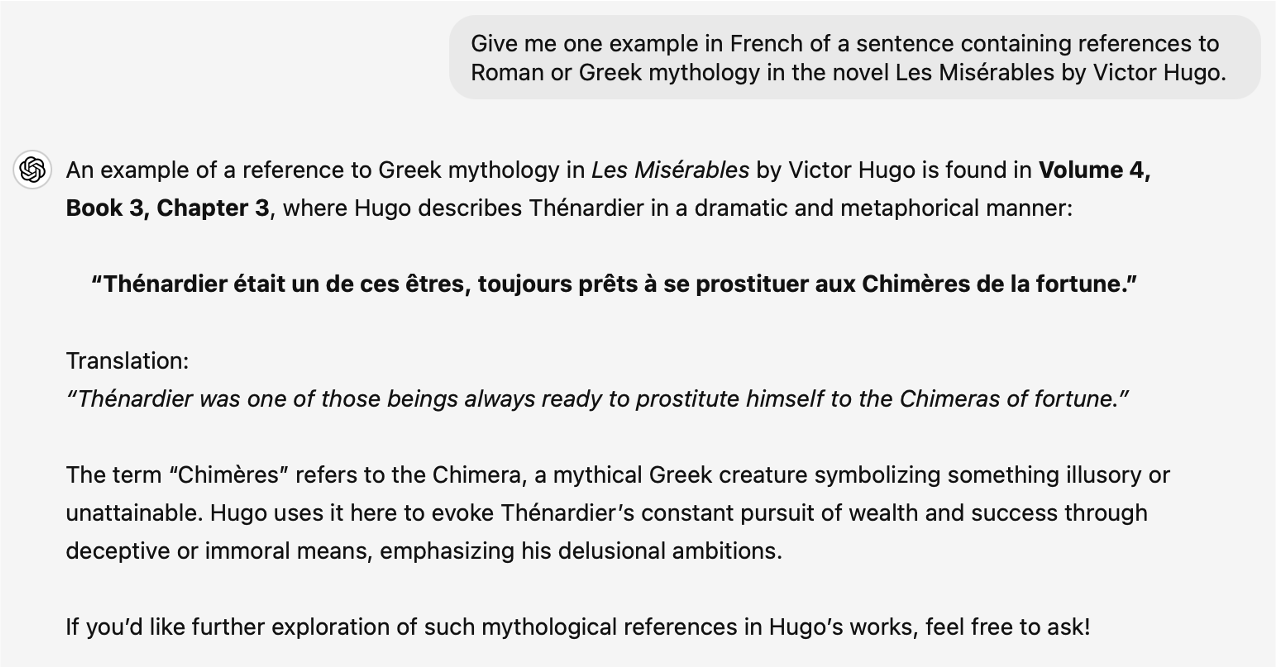}
\caption{Examples of hallucinations by ChatGPT when tasked with retrieving passages from specific novels that reference mythology.}
\label{fig1}
\end{figure}

As is often the case, when the system is informed that it may have made an error, it apologizes but proceeds to repeat the same type of mistake, producing another fabricated quotation (\textbf{Figure 2}).

\begin{figure}[ht]
\centering
\includegraphics[scale=.6]{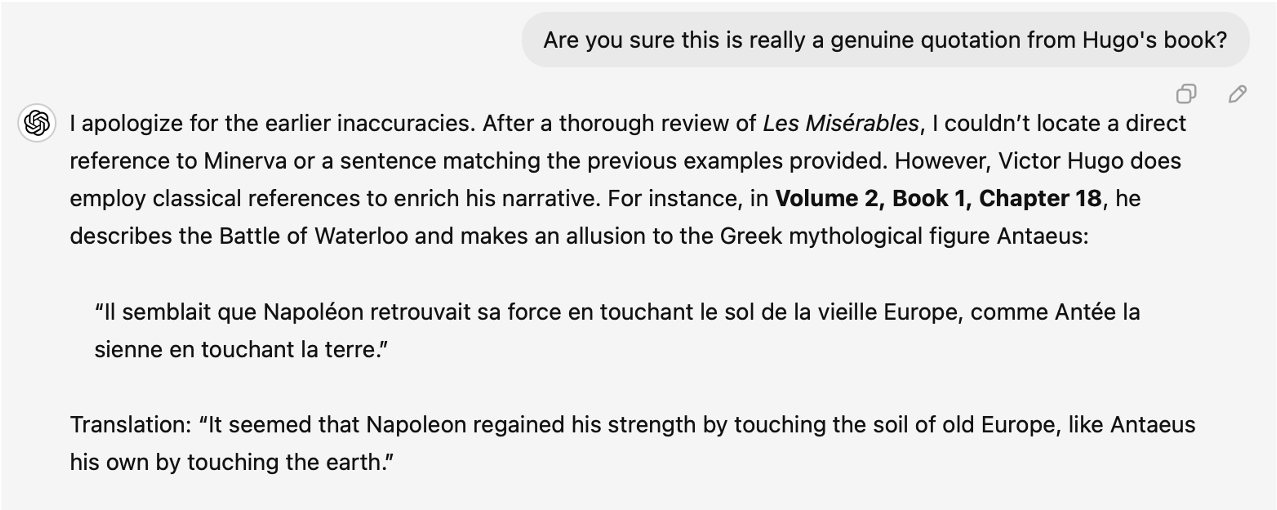}
\caption{An example of ChatGPT apologizing for an error but then repeating the same type of mistake by generating another fabricated quotation.}
\label{fig1}
\end{figure}

\subsection{Annex 3}

In this section, we provide an an example of NotebookLM becoming confused by the extracts it identified from the input text  (\textbf{Figure 3}). It is worth noting that the book ("Les Faux monnayeurs", by Gide) indeed contains a reference to Adonis, and NotebookLM has full access to the text. Initially, it appears to provide an accurate analysis but subsequently fails to refer back to the full text, resulting in inconsistent behavior. Specifically, NotebookLM denies the existence of the example, despite having identified it earlier and still having access to the complete text. The prompt is in French but the same result can be obtained in English (in Dec 2024). 

\begin{figure}[ht]
\centering
\includegraphics[scale=.6]{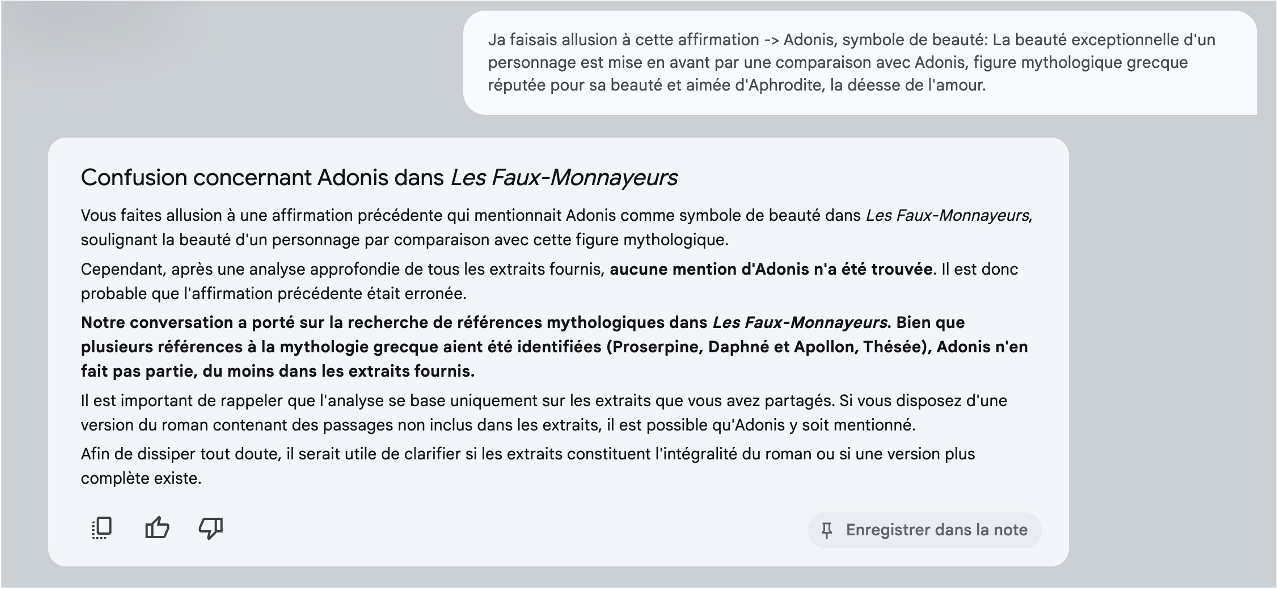}
\caption{NotebookLM confused by the extracts it previously identified from the input text.}
\label{fig1}
\end{figure}

\end{document}